\documentclass[10pt,twocolumn,letterpaper]{article}

\usepackage{iccv}
\usepackage{times}
\usepackage{epsfig}
\usepackage{graphicx}
\usepackage{amsmath}
\usepackage{amssymb}
\usepackage{amsthm}
\usepackage{booktabs}
\usepackage{tabularx}
\usepackage[accsupp]{axessibility}  % Improves PDF readability for those with disabilities.

\usepackage[breaklinks=true,bookmarks=false]{hyperref}

\iccvfinalcopy % *** Uncomment this line for the final submission

\def\iccvPaperID{2139} % *** Enter the ICCV Paper ID here

\ificcvfinal\fi

\begin{document}

%%%%%%%%% TITLE
\title{PR-RRN: Pairwise-Regularized Residual-Recursive Networks for \\ Non-rigid Structure-from-Motion}

\newcommand*\samethanks[1][\value{footnote}]{\footnotemark[#1]}
\author{Haitian Zeng\textsuperscript{1}, Yuchao Dai\textsuperscript{2}, Xin Yu\textsuperscript{3}, Xiaohan Wang\textsuperscript{1,3}, Yi Yang\textsuperscript{4}\thanks{Corresponding author.}\\
Baidu Research\textsuperscript{1}, Northwestern Polytechnical University\textsuperscript{2}, \\ University of Technology Sydney\textsuperscript{3}, Zhejiang University\textsuperscript{4}\\
{\tt\small zenghaitian@baidu.com; daiyuchao@gmail.com; xin.yu@uts.edu.au;} \\
{\tt\small xiaohan.wang-3@student.uts.edu.au; yee.i.yang@gmail.com}
% For a paper whose authors are all at the same institution,
% omit the following lines up until the closing ``}''.
% Additional authors and addresses can be added with ``\and'',
% just like the second author.
% To save space, use either the email address or home page, not both
}

\maketitle
% Remove page # from the first page of camera-ready.
\ificcvfinal\thispagestyle{empty}\fi

%%%%%%%%% ABSTRACT
\begin{abstract}
   We propose PR-RRN, a novel neural-network based method for Non-rigid Structure-from-Motion (NRSfM). PR-RRN consists of Residual-Recursive Networks (RRN) and two extra regularization losses. RRN is designed to effectively recover 3D shape and camera from 2D keypoints with novel residual-recursive structure. As NRSfM is a highly under-constrained problem, we propose two new pairwise regularization to further regularize the reconstruction. The Rigidity-based Pairwise Contrastive Loss regularizes the shape representation by encouraging higher similarity between the representations of high-rigidity pairs of frames than low-rigidity pairs. We propose minimum singular-value ratio to measure the pairwise rigidity. The Pairwise Consistency Loss enforces the reconstruction to be consistent when the estimated shapes and cameras are exchanged between pairs. Our approach achieves state-of-the-art performance on CMU MOCAP and PASCAL3D+ dataset.
\end{abstract}

%%%%%%%%% BODY TEXT
\section{Introduction}

The reconstruction of 3D object shapes and camera motions from 2D observations is an important problem in computer vision. When the object is rigid, this problem is defined as rigid Structure-from-Motion (SfM) and it can be solved reliably using existing methods like \cite{Tomasi1992Shape}. Non-Rigid Structure-from-Motion (NRSfM) relaxes the assumption of a rigid object in SfM to a deforming one, leading to a more general and challenging problem.

\begin{figure}[t]
\begin{center}
% \fbox{\rule{0pt}{2in} \rule{0.9\linewidth}{0pt}}
   \includegraphics[width=0.9\linewidth]{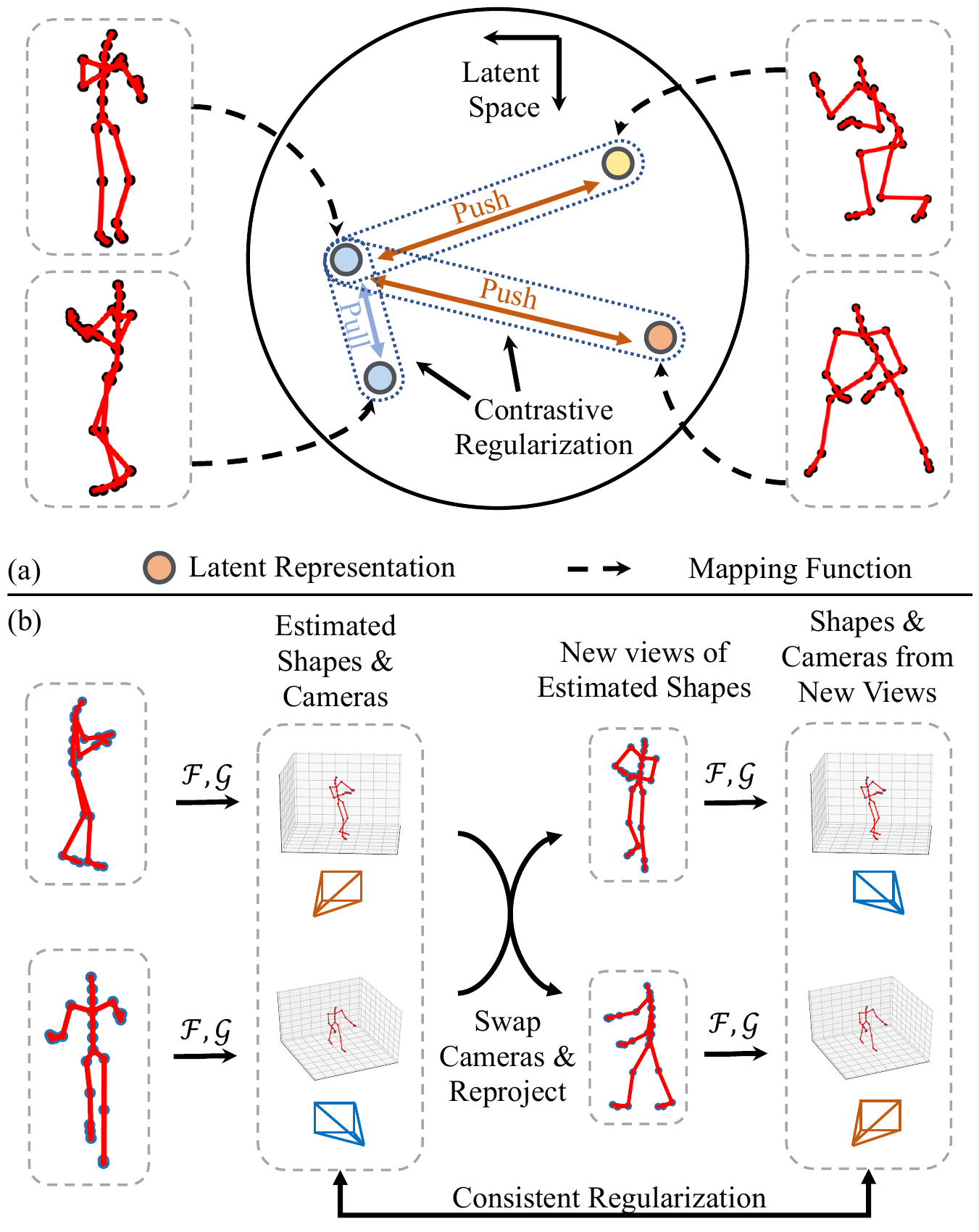}
\end{center}
   \caption{Illustration of pairwise losses. (a) Proposed pairwise contrastive regularization `pushes' or `pulls' the representations based on pairwise rigidity (similarity) of 2D shapes. (b) Consistent regularization forces the networks to produce consistent shape and camera estimation given new views of estimated shapes.}
\label{fig:fig1}
\end{figure}

NRSfM is known to be an under-constrained problem if the shape is allowed to deform arbitrarily in each observation. To make this problem tractable, a standard assumption is that in each frame the 3D shape is a linear combination of a small number of basis shapes \cite{DBLP:conf/cvpr/BreglerHB00}. With this assumption, NRSfM is formulated as factorizing the stacked observation matrix into three component matrices: \emph{cameras}, \emph{coefficients} and \emph{basis}. Previous researches exploit various constraints to solve this factorization problem, involving orthogonal constraint on the camera matrix \cite{Dai2012A, DBLP:conf/cvpr/XiaoK04}, restricting the basis to 3D shapes \cite{DBLP:journals/ijcv/XiaoCK06}. Different from those constraints on cameras or basis, another important category of approaches applied constraints to the coefficient matrix, including smooth trajectories over time in original coefficient matrix \cite{5728827,5639016} or in low-dimensional manifold \cite{6126319}, prior distributions \cite{7527684,4359359} and spatial smoothness \cite{DBLP:conf/eccv/HamsiciGM12}. In neural-network based models, the latent representation can be thought of as the `coefficients', and Sidhu \emph{et al.} \cite{DBLP:conf/eccv/SidhuTGAT20} first apply latent space constraints for sequential dense reconstruction. These constraints reduce the indeterminacy of the NRSfM task and potentially lead to better reconstruction.

% Novotny \emph{et al.} \cite{Novotny2020C3DPO} regularize the shape basis using transversal property.

However, regularizing the reconstruction is difficult when the data is large-scale and orderless. In such cases, assuming a representation manifold or using temporal smoothness is not possible. To tackle this, we propose to regularize the non-rigid shape reconstruction in a \emph{pairwise} manner. Compared to a strong global assumption of shapes, pairwise information are much easier to obtain, therefore the regularization can be achieved effectively.

% Inspired by recent advances in unsupervised representation learning \cite{DBLP:conf/cvpr/He0WXG20, DBLP:journals/corr/abs-1807-03748, DBLP:conf/eccv/TianKI20,yu2019unsupervised,shi2021self,DBLP:conf/eccv/Sanghi20}, where the model learns good representation by discriminating related pairs from unrelated pairs, we propose to regularize the non-rigid shape reconstruction in a \emph{pairwise} manner, as illustrated in Fig. \ref{fig:fig1}.
 
In this paper, we introduce Pairwise-Regularized Resi- dual-Recursive Networks (PR-RRN), a novel neural-network based model for NRSfM. PR-RRN consists of a Residual-Recursive Network (RRN) and two novel losses: Pairwise Contrastive Loss and Pairwise Consistency Loss. RRN alone can reconstruct the non-rigid shapes accurately, and it is further improved by pairwise losses. RRN contains a shape estimation network and a rotation estimation network, and the shape estimation network is constructed with a novel Residual-Recursive module, which is capable to enhance the reconstruction compared to a standard convolution layer with the same number of parameters. And the rotation estimation network is designed to estimate the camera matrix from the 2D input. Furthermore, two pairwise losses regularize the reconstruction in two different aspects, as shown in Fig.~\ref{fig:fig1}. Inspired by recent advances in unsupervised representation learning \cite{DBLP:conf/cvpr/He0WXG20, DBLP:journals/corr/abs-1807-03748, DBLP:conf/eccv/TianKI20,yu2019unsupervised,shi2021self,DBLP:conf/eccv/Sanghi20}, the proposed Pairwise Contrastive Loss encourages higher similarity between the latent representations of high-rigidity pairs of inputs than low-rigidity pairs. The pairwise rigidity is obtained by a novel measurement \emph{minimal singular-value ratio}. The Pairwise Consistency Loss enforces the reconstruction to be consistent when the estimated shapes and cameras are exchanged between pairs and reprojected as new inputs. The experimental results show that PR-RRN achieves state-of-the-art reconstruction performance on large-scale human motion and categorical objects datasets.

Our contributions are summarized as following:
\begin{itemize}
    \item We introduce a novel Residual-Recursive Network for non-rigid shapes reconstruction, which achieves state-of-the-art performance on CMU MOCAP Dataset.
    \item We propose Pairwise Contrastive Loss and Consistency Loss to further improve RNN. These two losses can regularize the reconstruction without assuming a global shape distribution.
    \item We design a novel pairwise rigidity measurement \emph{minimal singular-value ratio}. It is easy to compute and can be used to test the rigidity of a pair of 2D observations.
\end{itemize}

% (maybe related to group-based, graph-cluster?)

\begin{figure*}
\begin{center}
\includegraphics[width=1.0\linewidth]{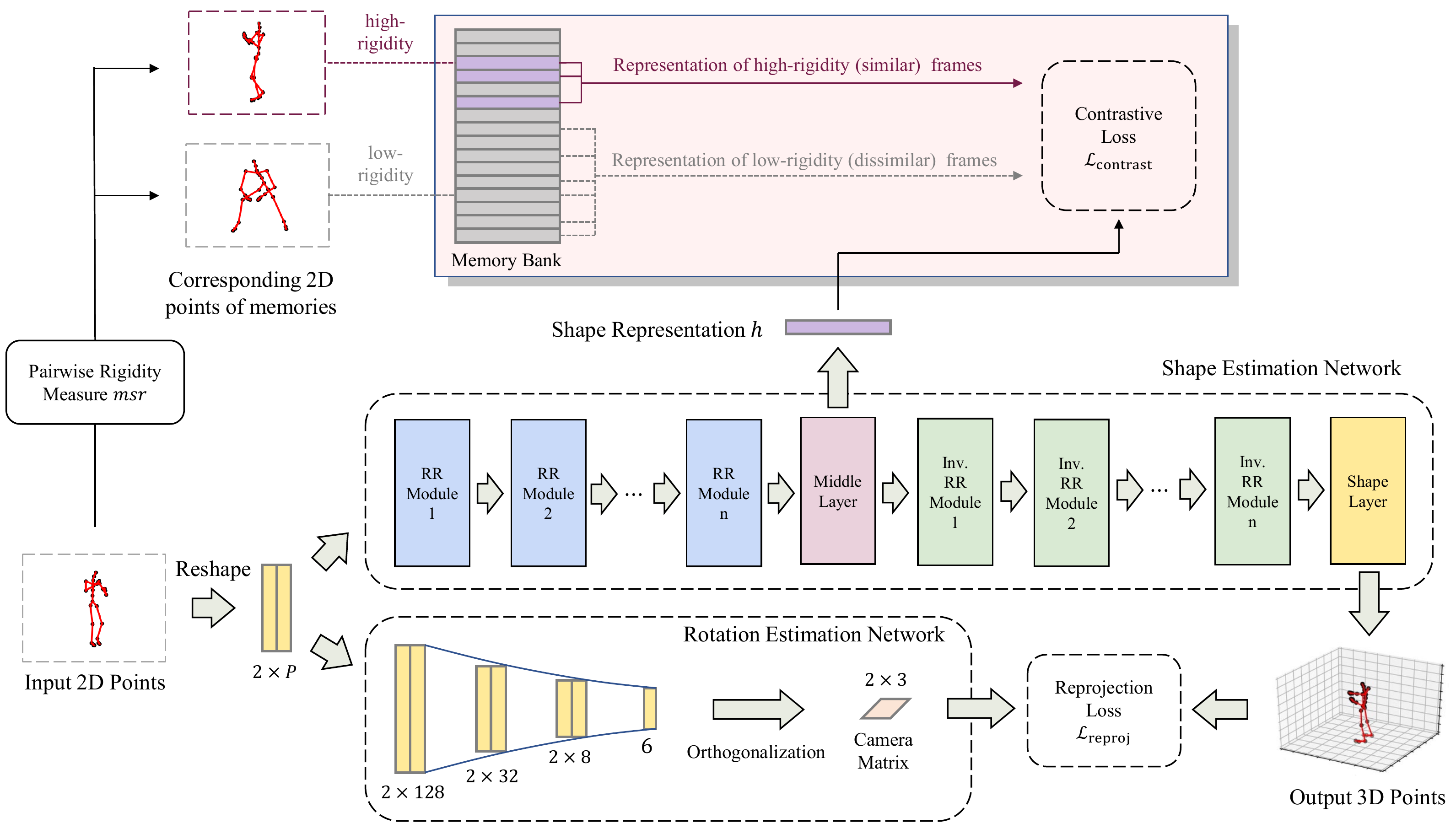}
\end{center}
   \caption{An overview of the proposed Residual-Recursive Networks and Pairwise Contrastive Loss. The RRN consists of two sub-networks: the shape estimation network and the rotation estimation network. The input 2D points are reshaped and fed into the two networks respectively. In the shape estimation network, a shape representation $h$ is produced. In the contrative loss, representations of other frames (memory) in the memory bank are divided into positive and negative examples using rigidity measure $\mathrm{msr}$, and are used to contrast with $h$. When the current training step is finished, $h$ will be stored in the memory bank, replacing the oldest memories.}
\label{fig:framework}
\end{figure*}

%------------------------------------------------------------------------
\section{Related Works}

\noindent\textbf{NRSfM.} The problem of non-rigid structure from motion is first introduced in the research of recovering a sequence of 3D face landmarks and camera positions by Bregler \emph{et al.} \cite{DBLP:conf/cvpr/BreglerHB00}. This research proposes a widely accepted assumption that the deforming 3D shapes can be compactly represented as a linear combination of a small number of basis shapes. Although this low rank assumption has been made, the deformation of shapes still remains under-constrained, which makes the NRSfM a challenging task for years. Various types of constraints \cite{DBLP:conf/nips/FragkiadakiSAM14,DBLP:conf/cvpr/XiaoK04,DBLP:conf/bmvc/FayadBAA09,DBLP:conf/eccv/VicenteA12} have been explored to restrict the deforming 3D structure. Xiao \emph{et al.} \cite{DBLP:journals/ijcv/XiaoCK06} propose a basis constraint to reach a close-form solution of NRSfM factorization. Torresani \emph{et al.} \cite{4359359} develop a Gaussian prior of the shape coefficients and reconstruct the shapes and cameras with probabilistic principal components analysis. With a sequence as the input, temporal smoothness can be leveraged to improve the reconstruction. Akhter \emph{et al.} \cite{5639016} introduce a dual representation of the NRSfM problem. 
% In their research, spatial trajectories of the points in the time-varying 3D shapes can be modeled with a low-rank DCT basis.
Gotardo \emph{et al.} \cite{5728827} formulate the temporal deformation of shapes as a smooth trajectory over the coefficients of shape basis, and this idea is improved by modeling the shape trajectory and basis shape in low-dimensional manifold \cite{DBLP:conf/iccv/ParkS11} and using kernel to measure the distance \cite{6126319}. There is a milestone that Dai \emph{et al.} \cite{Dai2012A} propose a block matrix method and achieve the outstanding performance with low rank priors. There are more breakthroughs \cite{DBLP:conf/cvpr/KumarCDL18, DBLP:journals/corr/KumarDL17, DBLP:conf/cvpr/Kumar19, Paladini2010SequentialNS, Agudo2020Unsupervised3R} in the field like inextensible \cite{DBLP:conf/cvpr/ChhatkuliPCB16, DBLP:conf/eccv/VicenteA12}, piecewise \cite{DBLP:conf/eccv/FayadAB10} methods, metric projection \cite{Paladini2009FactorizationFN}.

Further researches have been made to extend the NRSfM problem to more challenging situations. Zhu \emph{et al.} \cite{6909596} show that complex non-rigid human motions adhere to a union of subspace and solve it by a combined optimization of NRSfM and Low Rank Representation \cite{DBLP:conf/icml/LiuLY10}. Li \emph{et al.} \cite{DBLP:conf/cvpr/LiLJLS18} exploit grouped recurrent shapes and perform rigid SfM. Deep models have been applied to NRSfM \cite{DBLP:conf/eccv/ParkLK20,DBLP:conf/eccv/SidhuTGAT20}. A recent work of Kong \emph{et al.} \cite{9099404} proposes to solve the NRSfM problem by learning a multi-layer sparse dictionary which is approximated with a deep neural networks. Novotny \emph{et al.} \cite{Novotny2020C3DPO} introduce a factorization network and a canonicalization network to learn shape basis with transversal property. Sidhu \emph{et al.} \cite{DBLP:conf/eccv/SidhuTGAT20} build a deep model for sequential dense non-rigid shape reconstruction and show that the latent space constraints are useful.

% \textbf{3D Pose Estimation.} Our work is also related to 3D pose estimation, especially the cases without 3D ground truth supervision. Chen \emph{et al.} \cite{8953799} trained a lifting network to convert a 2D pose to 3D, and adopted an adversarial discriminator to distinguish the random projection of lifted 3D pose from real 2D poses. Tome \emph{et al.} \cite{8100086} proposed to simultaneously learn to estimate 2D landmark in images and to lift landmarks with a multi-stage network. Pavllo \emph{et al.} \cite{8954163} introduced a temporal convolutional network for 3D pose estimation and a semi-supervised learning scheme to benefit from extra unlabeled data. Wang \emph{et al.} \cite{8611195} adopted one 2D-to-3D network and one 3D-to-2D projection network to learn 3D human pose estimation from abundant images in a self-supervised manner with geometry consistency.

\noindent\textbf{Unsupervised Representation Learning.} Researches on unsupervised representation learning have achieved remarkable success. He \emph{et al.} \cite{DBLP:conf/cvpr/He0WXG20} propose Momentum Contrast (MoCo) for learning representations from unlabeled images, and the learned features are shown to be useful for downstream tasks. Oord \emph{et al.} \cite{DBLP:journals/corr/abs-1807-03748} present the noise contrastive learning with InfoNCE loss, and show that InfoNCE loss maximizes the lower bound of mutual information between related representations. Tian \emph{et al.} \cite{DBLP:conf/eccv/TianKI20} introduce a Contrastive Multiview Coding method for unsupervised learning of multi-view (or multi-modal) data by using the multiple views of a same example as positive pairs. Contrastive learning is also exploited for learning representations of 3D objects by Sanghi \cite{DBLP:conf/eccv/Sanghi20} \emph{et al.}, where the learned representation is shown to be useful for retrieving different views of a rigid object or similar objects.

% \textbf{Our Work.} Our work focus on the NRSfM problem. Compared to previous NRSfM works, the setting of our work is most close to Dai \emph{et al.} \cite{Dai2012A} where no prior knowledge of the shape or temporal continuity is exploited. We do not constraint the camera motion, different from \cite{5728827} where a coordinate system is set and smooth camera trajectories are assumed. Our work is capable to handle large scale data and complex variations of shapes, which were issued in \cite{6909596,9099404}. Our work on NRSfM is different from most 3D pose estimation works where extra supervisions like ground-truth 3D locations, anthropometric constraints and temporal smoothness are leveraged. 
%-------------------------------------------------------------------------

%------------------------------------------------------------------------
\section{NRSfM Recap}
We first briefly review the classic Non-Rigid Structure-from-Motion problem. The inputs of NRSfM problem are $F$ frames of $P$ keypoints, which are 2D views of a deformable object. Let the $i$-th frame to be $\mathrm{W}_i \in \mathbb{R}^{2 \times P}$, containing $P$ 2D coordinates. Under the condition of orthographic projection, the camera matrix of $i$-th frame is $\mathrm{M}_i \in \mathbb{R}^{2 \times 3}$, and satifies $\mathrm{M}_i \mathrm{M}_i^{\mathrm{T}} =\mathrm{I}_2$. The reconstructed 3D shape of $i$-th frame denotes $\mathrm{S}_i \in \mathbb{R}^{3 \times P}$, and it is related to $\mathrm{M}_i$ and $\mathrm{W}_i$ by the following projection equation:
\begin{eqnarray}
  \mathrm{W}_i = \mathrm{M}_i \mathrm{S}_i.
  \label{eq:proj_eq}
\end{eqnarray}

The NRSfM problem is known \cite{DBLP:conf/cvpr/XiaoK04} to be ill-posed if no assumption is made on $\mathrm{S}_i$. Bregler \emph{et al.} \cite{DBLP:conf/cvpr/BreglerHB00} make a widely-accepted assumption that $\mathrm{S}_i$ of all frames are a linear combination of $K$ basis shapes $\mathrm{B}_{k} \in \mathbb{R}^{3 \times P}$, which is:
\begin{eqnarray}
  \mathrm{S}_{i} = \sum_{k=1}^{K} (c_{i,k}\otimes \mathrm{I}_3)\mathrm{B}_k,
\end{eqnarray}
where $K\ll F$, $\otimes$ is the Kronecker product, and $c_{i,k}$ stands for the coefficient (weight) of $\mathrm{B}_k$ in $\mathrm{S}_{i}$.
% With this assumption, the factorization formulation of NRSfM is given as:
% \begin{eqnarray}
% \mathrm{W}^{\mathrm{mat}}
% =
% \mathrm{M}^{\mathrm{mat}}
% \left(
% \mathrm{C}^{\mathrm{mat}} \otimes \mathrm{I}_3
% \right)
% \mathrm{B}^{\mathrm{mat}}
% ,
% \end{eqnarray}
% where $\mathrm{W}^{\mathrm{mat}} \in \mathbb{R}^{2F \times P}$ is the stacked matrix of $\mathrm{W}_i$, $\mathrm{M}^{\mathrm{mat}} \in \mathbb{R}^{2F \times 3F}$ is the block diagonal matrix of all $\mathrm{M}_i$, $\mathrm{C}^{\mathrm{mat}} \in \mathbb{R}^{F \times K}$ is matrix of $c_{i,k}$, $\mathrm{B}^{\mathrm{mat}} \in \mathbb{R}^{3K \times P}$ is the stacked matrix of all $\mathrm{B}_k$.

In case of a rigid object, \emph{i.e.} the 3D shape does not deform across frames, NRSfM degrades into a Structure from Motion (SfM) problem, which can be formulated as:
\begin{eqnarray}
\mathbf{W}
=
\left[
 \begin{matrix}
  \mathrm{W}_{1} \\
  \vdots \\
  \mathrm{W}_{F} \\
 \end{matrix}
\right]
=
\left[
 \begin{matrix}
  \mathrm{M}_{1} \\
  \vdots \\
  \mathrm{M}_{F} \\
 \end{matrix}
\right]
\mathrm{S}^{r},
\end{eqnarray}
where $\mathbf{W} \in \mathbb{R}^{2F \times P}$ is the stacked matrix of $\mathrm{W}_i$, $\mathrm{S}^{r} \in \mathbb{R}^{3 \times P}$ is the rigid shape. So that in this case $\mathrm{rank}(\mathbf{W}) \leq 3$, and Tomasi \& Kanade \cite{Tomasi1992Shape} use a truncated Singular Value Decomposition (SVD) of $\mathbf{W}$ to recover the cameras and the rigid shape. This property is used in Sec.~\ref{sec:contrastloss} to derive a measure of rigidity.

\begin{figure}[t]
\begin{center}
% \fbox{\rule{0pt}{2in} \rule{0.9\linewidth}{0pt}}
   \includegraphics[width=1.0\linewidth]{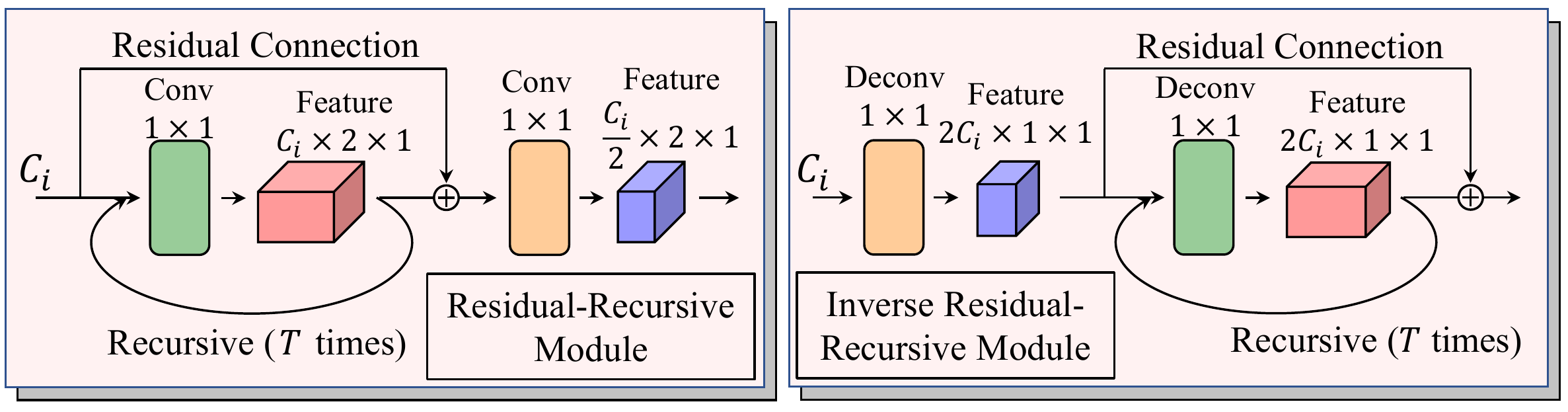}
\end{center}
   \caption{Design of the Residual-Recursive Module. Suppose the channel of input feature is $C_i$. The feature is repeatedly fed into the recursive layer for $T$ times before fed to the next layer.}
\label{fig:fig_RRM}
\end{figure}

\section{Method}
In this section, we introduce the Pairwise-Regularized Residual-Recursive Networks. The neural-network model is described in Sec.~\ref{sec:RRN} and the two pairwise-regularization losses are explained in Sec.~\ref{sec:contrastloss} and Sec.~\ref{sec:consistloss}. We suppose that the 2D keypoints in each frames are zero-centered so that the transition term is cancelled.

%-------------------------------------------------------------------------
\subsection{Residual-Recursive Networks}
\label{sec:RRN}
The Residual-Recursive Networks (RRN) consist of two sub-networks: the shape estimation network and the rotation estimation network. With a reprojection loss described in Sec.~\ref{sec:REN}, RRN can be trained to reconstruct 3D shapes from 2D keypoints.
\subsubsection{Shape estimation network}
\label{sec:SEN}
The role of shape estimation network is to map a single 2D input to a 3D shape. Let $\mathrm{W}_{i} \in \mathbb{R}^{2 \times P}$ be the $i$-th 2D input, the output of shape estimation network is the 3D shape $\mathrm{\hat{S}}_{i} \in \mathbb{R}^{3 \times P}$, which can be written as:
\begin{eqnarray}
\mathrm{\hat{S}}_{i} = \mathcal{F}(\mathrm{W}_{i}).
\end{eqnarray}

\noindent \textbf{Network Input.} The input $\mathrm{W}_{i}$ is reshaped into a $P \times 2 \times 1$ feature tensor which can be fed into convolution layers. Here, the number of channels is $P$, the width is $2$ and the height is $1$. We empirically find that it works better than vectorizing $\mathrm{W}_{i}$ into $2P \times 1 \times 1$ like in \cite{Novotny2020C3DPO, 8953799}. 

\noindent \textbf{Network Structure.} The shape estimation network is an autoencoder consisting of $n$ Residual-Recursive (RR) modules. The overall framework is shown in Fig.~\ref{fig:framework}. In each RR module, the feature is first fed into a recursive layer with residual connection, then it is processed with a fully connected layer where the number of output channels is reduced to half. The details of RR module is illustrated in Fig.~\ref{fig:fig_RRM}. After a middle layer, the feature is mapped to a 3D shape with Inverse Residual-Recursive modules. The two types of RR modules are illustrates in Fig.~\ref{fig:fig_RRM}. A tied-weight strategy \cite{9099404} is used here. Finally, the 3D shape is produced by a shape layer. Motivated by previous works of NRSfM where the compactness of model is emphasized~\cite{4359359, 6126319}, we choose the residual-recursive structure~\cite{zheng2020single,zhang2021recursive, DBLP:conf/iccv/DahlNS17} to enhance the representation power of a standard convolution layer without increasing the number of parameters. We empirically find that this structure is more effective than a standard convolution layer in an autoencoder for learning difficult 2D-3D mapping.
% The advantage of the residual-recursive design is that, the recursive layer makes the network deeper with fewer parameters compared to multiple normal layers \cite{DBLP:conf/iccv/DahlNS17}, and the residual connection avoids the vanishing gradient problem \cite{DBLP:conf/cvpr/HeZRS16}.

\noindent \textbf{Shape Representation.} We designate the output of the middle layer as the shape representation $h_i$. The regularization on the shape representation is explained in Sec.~\ref{sec:contrastloss}.

\subsubsection{Rotation estimation network}
\label{sec:REN}
In order to estimate the camera, we design a rotation estimation network to output an orthographic projection matrix $\mathrm{\hat{M}}_{i} \in \mathbb{R}^{2 \times 3}$ for a given 2D input, that is:
\begin{eqnarray}
\mathrm{\hat{M}}_{i} = \mathcal{G}(\mathrm{W}_{i}).
\end{eqnarray}

We suppose that the rotation is arbitrary in each frame. A recent work \cite{DBLP:conf/cvpr/ZhouBLYL19} shows that one needs 6D representation as well as a good mapping to avoid discontinuity while using the polar decomposition for enforcing orthogonality. The rotation estimation network is designed to output a 6D vector. The network is constructed with multiple linear layers, as shown in Fig.~\ref{fig:framework}. The output of the linear layers is reshaped into $\mathrm{\tilde{M}}_i \in \mathbb{R}^{2 \times 3}$.

The output $\mathrm{\tilde{M}}_i$ should be further turned into an orthographic projection matrix, \emph{i.e}. $\mathrm{\hat{M}}_{i}\mathrm{\hat{M}}_{i}^{\mathrm{T}}=\mathrm{I}_{2}$. The orthogonalization can be done in several ways, like Gram-Schmidt procedure \cite{DBLP:conf/cvpr/ZhouBLYL19}, projections onto $\mathrm{SO}(3)$ \cite{Paladini2009FactorizationFN, DBLP:conf/cvpr/KongL16}. We follow \cite{8953799, 9099404} to use Singular Value Decomposition (SVD) as the orthogonalization method, as it is shown that SVD produces better rotation estimation than Gram-Schmidt in many supervised and unsupervised tasks \cite{DBLP:conf/nips/LevinsonECSKRM20}. The orthogonalization process can be expressed as:
\begin{eqnarray}
\mathrm{\hat{M}}_{i}=\mathrm{UV^T} \quad \mathrm{s.t.} \quad  \mathrm{\tilde{M}}_i=\mathrm{U\Sigma V^T},
\end{eqnarray}
where $\mathrm{U\Sigma V^T}$ is the SVD of $\mathrm{\tilde{M}}_i$.

Finally, the 3D shape $\mathrm{\hat{S}}_{i}$ estimated by the shape estimation network is reprojected to a 2D shape using the rotation $\mathrm{\hat{M}}_{i}$. The reprojection loss is calculated as:
\begin{eqnarray}
\mathcal{L}_\textrm{reproj}=\left\| \mathrm{W}_i - \mathrm{\hat{M}}_{i}\mathrm{\hat{S}}_{i} \right\|_F,
\end{eqnarray}
where $\left\| \cdot \right\|_F$ is the Frobenius norm.

%-------------------------------------------------------------------------
\subsection{Rigidity-based Pairwise Contrastive Loss}
\label{sec:contrastloss}

We introduce a Rigidity-based Contrastive loss to improve the performance of deformable shape reconstruction. First, we define a new rigidity measure. Then, a Rigidity-based Contrastive loss is proposed.

\subsubsection{Minimal Singular Value Ratio}
Given two 2D frames, $\mathrm{W}_{i}$ and $\mathrm{W}_{j}$, we consider to measure the rigidity of them, \emph{i.e.}, how similar are the corresponding 3D shapes of the two frames. Note that $\mathrm{W}_{i}$ and $\mathrm{W}_{j}$ are randomly selected frames. Let $\mathrm{A} \in \mathbb{R}^{4 \times P}$ to be the stacked matrix of $\mathrm{W}_{i}$ and $\mathrm{W}_{j}$:
\begin{eqnarray}
\mathrm{A}=
\left[
 \begin{matrix}
  \mathrm{W}_{i} \\
  \mathrm{W}_{j} \\
 \end{matrix}
\right].
\end{eqnarray}
Inspired by Hamsici \emph{et al.} \cite{DBLP:conf/eccv/HamsiciGM12}, we use the ratio of the minimal (\emph{i.e.} fourth) singular value of $\mathrm{A}$ to define a novel rigidity measure $\mathrm{msr}$:
% \begin{definition}
%   This is a definition.
% \end{definition}
\begin{eqnarray}
\mathrm{msr}(\mathrm{W}_{i},\mathrm{W}_{j})=
\frac{\sigma_4^2}{\sum_{l=1}^{4}\sigma_l^2},
\end{eqnarray}
where $\sigma_l$ is the $l$-th singular value of $\mathrm{A}$ in descending order. As $\mathrm{rank}(\mathrm{A}) \leq 4$, the range of $\mathrm{msr}$ is $[0,0.25]$. Intuitively, $\mathrm{msr}$ measures how much $\mathrm{A}$ is away from a rank-3 matrix. If $\mathrm{rank}(\mathrm{A}) \leq 3$, then $\mathrm{msr}=0$ and it means that $\mathrm{W}_{i},\mathrm{W}_{j}$ are two views of a rigid 3D structure. On the contrary, the rigidity of $\mathrm{W}_{i},\mathrm{W}_{j}$ becomes lower when $\mathrm{msr}$ grows. More qualitative examples of $\mathrm{msr}$ are provided in Fig.~\ref{fig:tsne}.

\subsubsection{Rigidity-based Contrastive Loss}

We now introduce the Rigidity-based Contrastive Loss. This loss aims to regularize the representation of shapes by encouraging high similarity between similar shapes. The similarity of shapes can be found using the rigidity measure $\mathrm{msr}$ proposed previously. This regularization can be performed without assuming a global distribution or manifold of representation.

For a given frame $\mathrm{W}_{i}$, we calculate a \emph{positive set} $\mathcal{P}_i$ and a \emph{negative set} $\mathcal{N}_i$. The positive set contains the indices of frames that are (near) rigid with $\mathrm{W}_{i}$ measured by $\mathrm{msr}$, and \emph{vice versa}. These two sets are defined as:
\begin{align}
\mathcal{P}_i&=\{j|\mathrm{msr(\mathrm{W}_{i},\mathrm{W}_{j})<\tau,\forall{j}}\},\\
\mathcal{N}_i&=\{k|\mathrm{msr(\mathrm{W}_{i},\mathrm{W}_{k})>\xi,\forall{k}}\},
\end{align}
where $\tau,\xi$ are threshold parameters. 
\begin{table*}
\centering
\begin{tabular}[t]{l|c|c|c|c|c|c|c|c|c}
\noalign{\hrule height 1.0pt}
\vspace{1.0pt}
Methods                & Subj. 07    & Subj. 20    & Subj. 23    & Subj. 33    & Subj. 34    & Subj. 38    & Subj. 39    & Subj. 43  & Subj. 93 \\
% \noalign{\hrule height 1.0pt}
\hline
CSF \cite{5728827}                   & 1.231 & 1.164 & 1.238 & 1.156 & 1.165 & 1.188 &  1.172 & 1.267 & 1.117 \\
URN \cite{9010628} & 1.504 & 1.770 & 1.329 & 1.205 & 1.305 & 1.303 & 1.550 & 1.434 & 1.601 \\
CNS \cite{8778692}                  & 0.310 & 0.217 & 0.184 & 0.177 & 0.249 & 0.223 & 0.312 & 0.266 & 0.245 \\
C3DPO \cite{Novotny2020C3DPO}        & 0.226 & 0.235 & 0.342 & 0.357 & 0.354 & 0.391 & 0.189 & 0.351 & 0.246 \\
DNRSFM \cite{9099404}                & 0.045  & 0.137 & 0.053 & 0.137 & 0.062 & 0.053 & 0.041 & 0.125 & 0.214 \\
% \noalign{\hrule height 1.0pt}
\hline
PR-RRN (Ours)             & \textbf{0.024} & \textbf{0.034} & \textbf{0.039} & \textbf{0.043} & \textbf{0.039} & \textbf{0.034} & \textbf{0.025} & \textbf{0.028} & \textbf{0.152} \\
\hline
PR-RRN (Unseen)               & 0.061 & 0.167 & 0.249 & 0.254 & 0.265 & 0.108 & 0.028 & 0.080 & 0.242 \\

\noalign{\hrule height 1.0pt}
\end{tabular}
\vspace{0.25em}
\caption{The reconstruction error $\mathrm{e_{3D}}$ on CMU MOCAP.}
\label{tab:cmu_mocap}
\end{table*}

The Rigidity-based Contrastive Loss is:
\begin{align}
\mathcal{L}_\mathrm{contrast}=-\mathbb{E}
  \left[
\mathrm{log}\frac{\sum_{j}{\mathrm{exp}(h_i \cdot h_j})}{\sum_{j}{\mathrm{exp}(h_i \cdot h_j}) + \sum_{k}{\mathrm{exp}(h_i \cdot h_k})}
\right],
\end{align}
where $\cdot$ is the dot product, $j \in \mathcal{P}_i$ and $k \in \mathcal{N}_i$. Intuitively, this loss is minimized when $h_i \cdot h_j$ have high values and $h_i \cdot h_k$ have low values. We normalize all $h$ to unit norm before calculating the loss.

In practice, the networks are trained with mini-batches, therefore the frames outside the current mini-batch are not available. To deal with that, we use a memory bank \cite{DBLP:conf/cvpr/He0WXG20} to store representations from previous mini-batches. The size of the memory bank is $N_{mem}$. After each training step, the current batch of representations are stored in memory bank, replacing the oldest ones. In other words, the memory bank works as a queue of representation. This allows the representation to be regularized by as much pairs as possible, and it is proved to be beneficial to learning good representation \cite{ DBLP:journals/corr/abs-1807-03748, DBLP:conf/cvpr/He0WXG20}.

\subsection{Pairwise Consistency Loss}
\label{sec:consistloss}

In this subsection we propose a novel pairwise consistency constraint. Given two random 2D frames $\mathrm{W}_{i},\mathrm{W}_{j}$, the shapes and rotations are estimated using the $\mathcal{F}$ and $\mathcal{G}$ of RRN:
\begin{align}
\mathrm{\hat{S}}_i &= \mathcal{F}(\mathrm{W}_i), \quad \mathrm{\hat{M}}_i = \mathcal{G}(\mathrm{W}_i), \\
\mathrm{\hat{S}}_j &= \mathcal{F}(\mathrm{W}_j), \quad \mathrm{\hat{M}}_j = \mathcal{G}(\mathrm{W}_j).
\end{align}

If the positions of estimated camera motion $\mathrm{\hat{M}}_i$, $\mathrm{\hat{M}}_j$ are exchanged and further reprojected with $\mathrm{\hat{S}}_i$, $\mathrm{\hat{S}}_j$, two new observations $\mathrm{W}_i^{'}$, $\mathrm{W}_j^{'}$ can be obtained, that is:
\begin{eqnarray}
  \mathrm{W}_i^{'} = \mathrm{\hat{M}}_j \mathrm{\hat{S}}_i, \quad
  \mathrm{W}_j^{'} = \mathrm{\hat{M}}_i \mathrm{\hat{S}}_j.
\end{eqnarray}
The proposed Pairwise Consistency enforces $\mathcal{F}$ and $\mathcal{G}$ to estimate $\mathrm{W}_i^{'}$, $\mathrm{W}_j^{'}$ consistently back to $[\mathrm{\hat{M}}_i,\mathrm{\hat{S}}_j]$.

This idea can be easily extended from two frames to a mini-batch of $L$ frames. Given the output $\{\mathrm{\hat{M}}_i,\mathrm{\hat{S}}_i\}_L$ of $\mathcal{G},\mathcal{F}$, we produce a new batch of 2D observations $\{\mathrm{W}_i^{'}\}_L$ by performing the following reprojections:
\begin{eqnarray}
  \mathrm{W}_i^{'} = \mathrm{\hat{M}}_{r_i} \mathrm{\hat{S}}_i,
\end{eqnarray}
where $r_1 \dots r_L$ is a random permutation of $1 \dots L$. In order to enforce the Pairwise Consistency, the Pairwise Consistency Loss $\mathcal{L}_\textrm{consist}$ is applied to the training process of the model, and it is calculated as:
\begin{eqnarray}
  \mathcal{L}_\textrm{consist} = \sum_{i=1}^{L}
  \left\|
  \mathrm{\hat{S}}_{i}-\mathrm{\hat{S}}_{i}^{'}
  \right\|_F
  +
  \left\|
  \mathrm{\hat{M}}_{i}-\mathrm{\hat{M}}_{i}^{'}
  \right\|_F,
\label{eq:Lcon}
\end{eqnarray}
where $\mathrm{\hat{S}}_{i}^{'}=\mathcal{F}(\mathrm{W}_i^{'})$, $\mathrm{\hat{M}}_{r_i}^{'}=\mathcal{G}(\mathrm{W}_i^{'})$, and  $\{\mathrm{\hat{M}}_{r_i}^{'}\}$ is rearranged to the original order $\{\mathrm{\hat{M}}_i^{'}\}$ using the inverse permutation of $r_1 \cdots r_L$. We notice that there are better measurements of rotation distance than the second term in \eqref{eq:Lcon}, but we empirically find that the Frobenius norm is also feasible and simple to implement. In addition, we find that replacing $\mathrm{\hat{M}}_i$ with a random rotation matrix is a good alternative which can slightly improve the performance.

% Note that in the training of deep networks, the dataset is randomly shuffled at the start of each epoch, so that the consistency loss is calculated for $L$ random frames of the datasets at each training step.

%-------------------------------------------------------------------------
\subsection{Alternative Training}
The final training objective of PR-RRN is:
\begin{eqnarray}
\mathcal{L}=\mathcal{L}_\textrm{reproj}+\lambda_1\mathcal{L}_\textrm{contrast}+\lambda_2\mathcal{L}_\textrm{consist},
\end{eqnarray}
where $\lambda_1$ and $\lambda_2$ are weighting parameters. We empirically find that training the networks using $\mathcal{L}_\textrm{contrast}$ and $\mathcal{L}_\textrm{consist}$ alternatively produces better results than using them jointly, while $\mathcal{L}_\textrm{reproj}$ is always used.

\begin{table*}
\centering
% \fontsize{14}{13}
% \selectfont
% \begin{tabular}{@{}l@{}|@{}l@{}|@{}l@{}|@{}l@{}|@{}l@{}|@{}l@{}|@{}l@{}|@{}l@{}|@{}l@{}|@{}l@{}|@{}l@{}|@{}l@{}|@{}l@{}|@{}l@{}}
\small
\setlength{\tabcolsep}{2.25mm}{  % just table width for fitting to the page width
\begin{tabular}{l|l|l|l|l|l|l|l|l|l|l|l|l|l}
% \hline
\noalign{\hrule height 1.0pt}
\vspace{1.0pt}
            & CSF   & KSTA  & BMM   & CNS   & NLO   & RIKS  & SPS   & SFC   & MUS   & URN & C3D & DNR & Ours \\ \hline
Aeroplane   & 0.363 & 0.175 & 1.459 & 0.416 & 0.876 & 0.132 & 0.930 & 0.504 & 0.261 & 0.121 & 0.272 & \textbf{0.024}  & 0.031 \\ \hline
Bicycle     & 0.424 & 0.245 & 1.376 & 0.356 & 0.269 & 0.136 & 1.322 & 0.372 & 0.178 & 0.328 & 0.585 & \textbf{0.003}  & 0.005 \\ \hline
Bus         & 0.217 & 0.199 & 1.023 & 0.250 & 0.140 & 0.160 & 0.604 & 0.251 & 0.113 & 0.097 & 0.271 & \textbf{0.004}  & 0.008 \\ \hline
Car         & 0.195 & 0.186 & 1.278 & 0.258 & 0.104 & 0.097 & 0.872 & 0.282 & 0.078 & 0.104 & 0.276 & 0.009  & \textbf{0.005} \\ \hline
Chair       & 0.398 & 0.399 & 1.297 & 0.170 & 0.146 & 0.192 & 1.046 & 0.226 & 0.210 & 0.115 & 0.658 & \textbf{0.007}  & 0.025 \\ \hline
Diningtable & 0.406 & 0.267 & 1.000  & 0.170 & 0.109 & 0.207 & 1.050 & 0.221 & 0.264 & 0.115 & 0.441 & 0.060  & \textbf{0.015} \\ \hline
Motorbike   & 0.278 & 0.255 & 0.857 & 0.457 & 0.432 & 0.118 & 0.986 & 0.361 & 0.222 & 0.287 & 0.492 & \textbf{0.002}  & 0.006 \\ \hline
Sofa        & 0.409 & 0.307 & 1.126 & 0.250 & 0.149 & 0.228 & 1.328 & 0.302 & 0.167 & 0.181 & 0.343 & \textbf{0.004}  & 0.007 \\ \hline
Average     & 0.336 & 0.223 & 1.178 & 0.291 & 0.278 & 0.159 & 1.017 & 0.315 & 0.186 & 0.168 & 0.417 & 0.014  & \textbf{0.013} \\ \hline
\noalign{\hrule height 1.0pt}
\end{tabular}
}
\vspace{1.0pt}
\caption{The reconstruction error $\mathrm{e_{3D}}$ on PASCAL3D+. The performances of compared methods are quoted from \cite{DBLP:conf/cvpr/AgudoPM18, 8953799, 9099404}.}
\label{Tab:PASCAL}
\end{table*}

%------------------------------------------------------------------------
\section{Experiments}
We evaluate our method on a large-scale human motion dataset, a categorical objects dataset, a facial landmark dataset and a mesh dataset, which are representative deformable shapes. We first introduce the datasets and the experimental setups. Next the reconstruction results are reported. Finally, we analyze the proposed model in detail.

\subsection{Datasets and Setups}
\label{subsec:datasets}
\textbf{CMU MOCAP.} The CMU Motion Capture dataset\footnote {http://mocap.cs.cmu.edu/} consists of 144 subjects, and most subjects contain tens of human activity sequences. In each activity, ground truth 3D coordinates of 31 keypoints are recorded in the world coordinate system. CMU MOCAP is diverse and large enough for verifying the PR-RRN. We select 9 subjects from CMU MOCAP. For fair comparison with previous methods, we build the training and testing set following \cite{9099404}: the first 80\% the activity sequences in a subject are concatenated as the training set and the remaining 20\% are used as testing set (Unseen). Random orthogonal projections are applied to 3D shapes to obtain 2D observations. The coordinates of the 3D shapes are centered to zero in each frame to cancel the transition term in camera projection. Note that in training deep networks, the data will be shuffled every epoch, so that the input frames are \emph{orderless}.

\textbf{PASCAL3D+.} PASCAL3D+ datasets \cite{DBLP:conf/wacv/XiangMS14} contains 12 categories of objects with 3D annotations from around 80 CAD models. Each category contains about 3000 objects on average. For fair comparison with previous works, we follow \cite{9099404} to use the categories with at least 8 points of annotation, and do not split the dataset into training and testing set. The ground truth 3D shapes and 2D observations are also zero-centered.

\textbf{MUCT Face.} The MUCT Face dataset \cite{Milborrow10} consists of 3755 faces with 76 facial landmarks annotations. The dataset is diverse in lighting, age and races. The face images are collected with five cameras from different viewpoints. In our experiments, we use all the 76 keypoints. As there is no 3D ground-truth of the points, we use the MUCT for qualitative evaluation.

\textbf{TWO CLOTHS.} The TWO CLOTHS dataset \cite{Taylor2010NonrigidSF} is a popular dataset for mesh reconstruction, which contains 163 frames of two fast deforming cloths. The dataset provides the 2D trajectory of a 525-point grid mesh.

\textbf{Handling Missing Points.} MUCT contains some missing points caused by occlusion. In the experiment on MUCT dataset, the input coordinates of missing 2D points are simply set to zero, and the normal points are subtracted by their mean to become zero-centered. In training, the $\mathrm{msr}$ are calculated only with common visible points of two observations, and the losses are masked with the visibility.

\textbf{Evaluation Metrics.} Following previous works \cite{8778692,9010628}, we use the normalized mean 3D error to evaluate the shape recovery accuracy. Before evaluation, the predicted 3D shape is aligned to the ground truth using Procrustes algorithm. The metric is calculated as:
\begin{eqnarray}
  e_\textrm{3D}=\frac{1}{F}\sum^{F}_{i=1}
  \frac{\left\|
  \mathrm{S}_{i}^{\mathrm{gt}}-\mathrm{\hat{S}}_{i}
  \right\|_2}
  {\left\|
  \mathrm{S}_{i}^{\mathrm{gt}}
  \right\|_2},
\end{eqnarray}
where $\mathrm{S}_{i}^{\mathrm{gt}}$ is the ground-truth 3D shape of $i$-th frame.

\textbf{Training Details.}
In all experiments, we set $\lambda_1=0.1$, $\lambda_2=0.2$, $\tau=0.02$, $\xi=0.04$, $N_{mem}=1024$. For the RRN, the number of RR modules $n$ is set to 5, the channels of the modules are $128, 64, 32, 16, 8$, so that the dimension of shape representation $h_i$ is $8$. In the rotation estimation network, the sizes of linear layers are $128, 32, 8$. The recursive time $T$ is set to $3$ for CMU MOCAP dataset and $2$ for PASCAL3D+. The network is trained with Adam optimizer with a learning rate of $0.001$ and an exponential decay rate of $0.95$ for $700$ epochs. The pairwise consistency loss and the contrastive loss are used alternatively for 100 epochs.

% For CMU MOCAP dataset, we set the number of cluster $N=16$ in all subjects. We trained two variants of the full model. One is the model only with clustered representation, denotes \textbf{Ours-CLS}. The other is model with clustered representation and similarity prediction, marked as \textbf{Ours-CLS-SIM}. For all subjects in CMU MOCAP and all variants models, we use $\lambda_1=0.1$, $\lambda_2=0.2$, $\lambda_3=0.05$, $t=0.3$ and $\tau=0.8$. The network is trained with Adam optimizer with a learning rate of $0.001$ and an exponential decay rate of $0.95$. To stabilize the training, we follow \cite{DBLP:conf/nips/JiZLSR17} to pretrain the backbone for 50 epochs.
% In the training process, we notice that models with the similarity prediction and decomposition consistency takes additional 200 epochs to converge

\subsection{NRSfM Results}
\textbf{CMU MOCAP.} PR-RRN is compared with several strong methods. As the number of frames is large, classic methods like \cite{6126319, Dai2012A, DBLP:conf/eccv/HamsiciGM12} fail, except CSF \cite{5728827} and CNS \cite{8778692}. CSF and CNS assume temporal smooth trajectories of points, so that the sequential frames of CMU MOCAP datasets will put these two methods in advantage. URN \cite{9010628}, C3DPO \cite{Novotny2020C3DPO}, DNRSFM \cite{9099404} and ours PR-RRN are deep models which can deal with large-scale reconstruction and do not assume temporal smoothness. Tab.~\ref{tab:cmu_mocap} shows the results on the 9 subjects of CMU MOCAP. For PR-RRN, the results of unseen shapes (test set) are reported. One can see that PR-RRN outperforms all four competing methods in 9 subjects. On Subject 20, 33 and 43, PR-RRN surpasses the state-of-the-art approaches by a large margin. It is also worth noting that PR-RRN achieves high accuracy when tested with Unseen shapes on Subject 07, 38, 39 and 43. This may be from a small domain gap between the training set and the testing set. In short, the results validate the capability of PR-RRN for accurate recovery of non-rigid shapes.

\textbf{PASCAL3D+.} For PASCAL3D+ we consider more methods for comparison, including CSF \cite{5728827}, KSTA \cite{6126319}, BMM \cite{Dai2012A}, CNS \cite{8778692}, NLO \cite{DBLP:journals/ivc/BueSA07}, RIKS \cite{DBLP:conf/eccv/HamsiciGM12}, SPS \cite{DBLP:conf/cvpr/KongL16}, SFC \cite{DBLP:conf/3dim/KongZKL16}, MUS \cite{DBLP:conf/cvpr/AgudoPM18}, URN \cite{8953799}, C3DPO \cite{Novotny2020C3DPO} and DNRSFM \cite{9099404}. The results are shown in Tab.~\ref{Tab:PASCAL}. PR-RRN and DNRSFM both achieve higher accuracy on all 8 selected categories of PASCAL3D+ than other methods, while the PR-RRN performs better than DNR on average and especially on \texttt{Diningtable} class. The results show that PR-RRN also performs well on categorical objects reconstruction tasks.

\textbf{MUCT.} The reconstruction results on MUCT dataset are visualized in Fig.~\ref{fig:MUCT}. From the results one can verify that the recovery of non-rigid facial landmarks is successful under realistic camera motions of MUCT dataset.

\textbf{TWO CLOTHS.} We test our method on the TWO CLOTHS dataset to validate mesh reconstruction. As there is no ground-truth, we visualize the qualitative result in Fig.~\ref{fig:twocloths}, where our model produces plausible deformation and clear segmentation of the two cloths.

\subsection{Model Analysis}
\textbf{Structure of RRN.} We give an ablation study to validate the residual-recursive design in RRN. We set up a \textbf{Vanilla} baseline where the shape network $\mathcal{F}$ contains standard convolution layers with same number of parameters as RRN, and the rotation network $\mathcal{G}$ remains the same as RRN. Note that the only difference between Vanilla and RRN is the residual-recursive structure. We compare the two models together with DNRSFM in Tab.~\ref{tab:ablation}. As shown, in Subject 20, 23, 33 and 43, RRN outperforms Vanilla by a margin, which verifies the effectiveness of the structure. The RRN structure does not work well on the difficult Subject 93, however it can be improved by the pairwise regularizations. It is also worth noting that Vanilla model outperforms DNRSFM in Subject 33 and 43 and is competitive in Subject 20 and 93.

\textbf{Effectiveness of Constrast and Consistent Losses.} To understand the effectiveness of proposed Residual-Recursive Networks and two novel losses, we conduct experiments with three variations of the PR-RRN: 1) \textbf{RRN.} The Residual-Recursive Networks trained with reprojection loss only. 2) \textbf{RRN-Contrast.} The RRN trained with reprojection loss and Pairwise Constrastive Loss. 3) \textbf{RRN-Consist.} The RRN trained with reprojection loss and Pairwise Consistency Loss. The reconstruction results are reported in Tab.~\ref{tab:ablation2}, together with the full model PR-RRN for comparison. From the table, one can see that the proposed RRN achieves high accuracy on CMU MOCAP Subject 20, 23, 33 and 43, and it is further improved by Contrast Loss and Consistency Loss. For Subject 93, the performance of RRN is significantly enhanced by the regularization losses.

\textbf{Limitations.} In our experiments, our method can address points from 8 to more than 500. However, the SVD in Contrastive Loss becomes a bottleneck for the entire model when handling a large scale of points, \emph{e.g.} 5000 points. When a shape contains this amount of points, training PR-RRN will become computationally prohibitive.

\begin{figure}[t]
% \vspace{-1.5em}
\begin{center}
\includegraphics[width=0.99\linewidth]{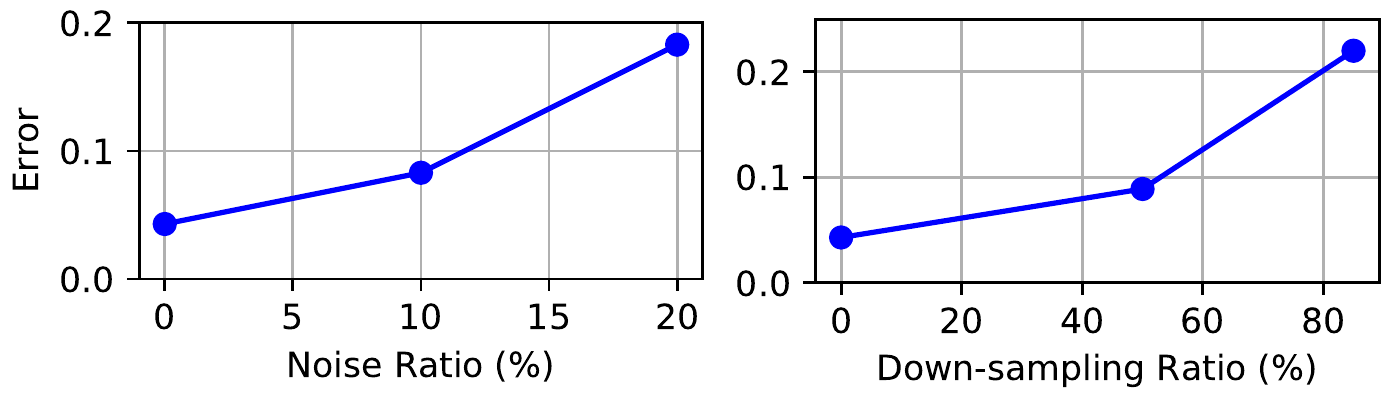}
\end{center}
\vspace{-1em}
\caption{Performance with noisy or down-sampled data.}
\label{fig:robust}
\end{figure}

\begin{figure}[t]
% \vspace{-1.5em}
\begin{center}
\includegraphics[width=0.99\linewidth]{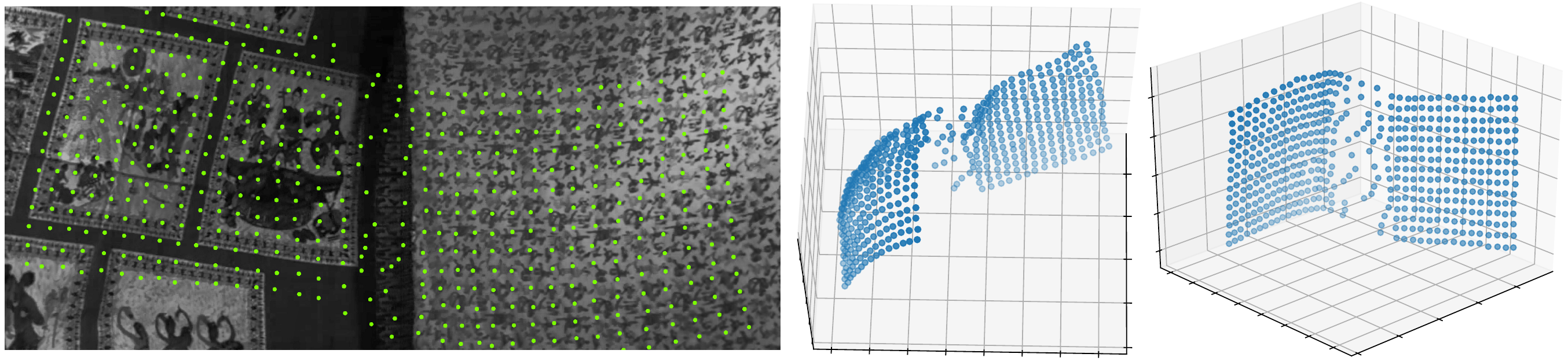}
\end{center}
\vspace{-1em}
\caption{Results on TWO CLOTHS Dataset.}
\label{fig:twocloths}
\end{figure}

\begin{figure}
\begin{center}
\includegraphics[width=1.0\linewidth]{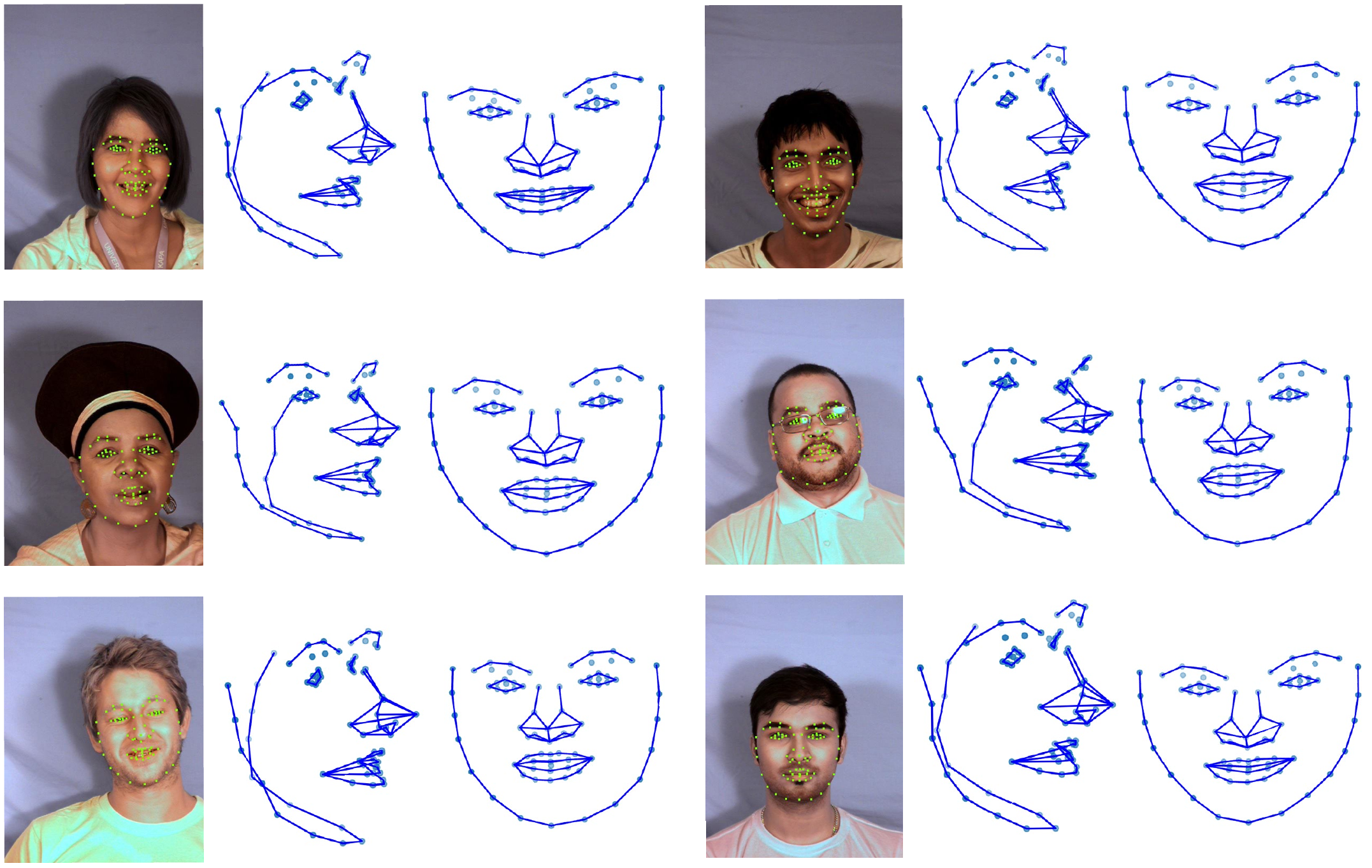}
\end{center}
\caption{Visualization of some reconstruction results on MUCT dataset \cite{Milborrow10}. Left: Origin pictures of different people. Center: Side views of the reconstructed shapes. Right: Front views of reconstructions.}
\label{fig:MUCT} % Must after caption.
\end{figure}

\begin{figure}
\begin{center}
\includegraphics[width=0.99\linewidth]{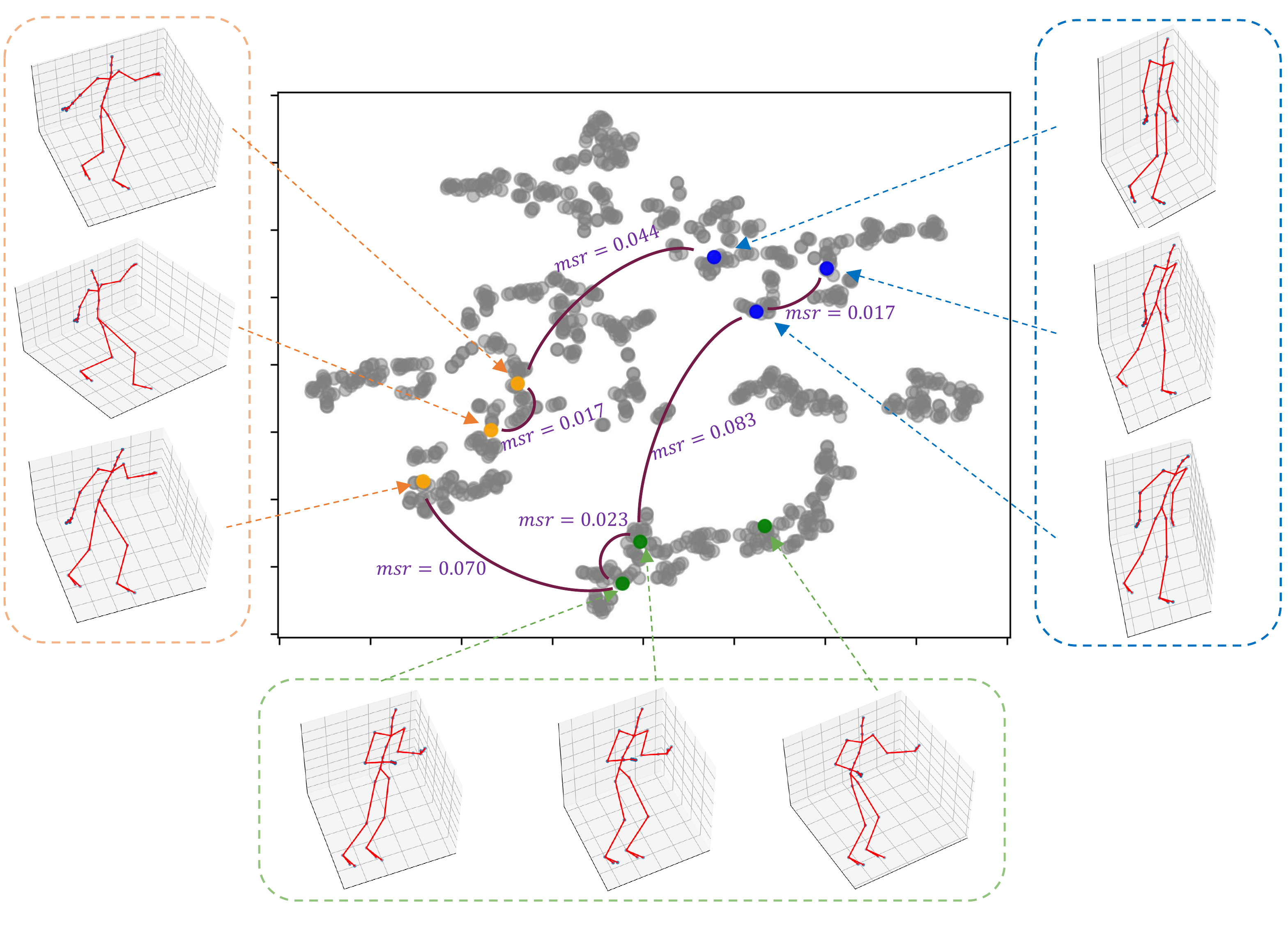}
\end{center}
\caption{t-SNE \cite{tsne} visualization of the shape representation learned by PR-RRN on CMU MOCAP Subject 20. The grey points are 1000 randomly selected frames out of a total of 4183 frames. We show 9 reconstructed shapes which can be coarsely divided into three groups. One can see that shape representations are spatially closer to the shapes in the same group than shapes in other groups. Additionally, we mark out some pairwise rigidity measure $\mathrm{msr}$, colored in purple. Qualitatively, the $\mathrm{msr}$ correctly reflects the similarity of different 3D shapes, and generally agrees with the distance of representation. Best viewed in color.}
\label{fig:tsne}
\end{figure}

\textbf{Robustness.} We analyze the robustness of PR-RRN under noisy and small-scale data. (1) We add Gaussian noise to Subject 33 of CMU MOCAP dataset. We follow \cite{9099404} to calculate the noise ratio: $\left\|\text{noise} \right\|_F / \left\|\mathbf{W} \right\|_F$. (2) We train our model on down-sampled Subject 33 and test it on full dataset. In Fig.~\ref{fig:robust}, one can see that the proposed method is capable to achieve reasonable accuracy on corrupted data.

\begin{table}[htb]
\small
\centering
\begin{tabular}[t]{lccccc}
\toprule
Model & 20 & 23 & 33 & 43 & 93\\
\midrule
DNRSFM \cite{9099404}  & 0.137  & 0.053  & 0.137 & 0.125 & 0.214 \\
\midrule
Vanilla & 0.147 & 0.352 & 0.060 & 0.072  & 0.213 \\
% Vanilla-Deep &   &    &    &   \\
% Vanilla-2x &   &    &    &   \\
% Vanilla-Res & 0.319  & 0.056  &  0.091  &   \\
% Vanilla-Recur & 0.077  &    &    &   \\
RRN & 0.041 & 0.050 & 0.051  & 0.047  & 0.305 \\
\bottomrule
\end{tabular}
\vspace{0.8em}
\caption{Analysis on RRN structure.}
\label{tab:ablation}
\end{table}

\begin{table}[htb]
\small
\centering
\begin{tabular}[t]{lccccc}
\toprule
Model & 20 & 23 & 33 & 43 & 93\\
\midrule
RRN & 0.041 & 0.050 & 0.051  & 0.047  & 0.305 \\
RRN-Contrast & 0.039 & 0.043 & 0.046  & 0.033  & 0.255 \\
RRN-Consist & 0.038 & 0.045 & 0.044  & 0.034  & 0.160 \\
\midrule
PR-RRN (full) & 0.034 & 0.039 & 0.043  & 0.028  & 0.152 \\
\bottomrule
\end{tabular}
\vspace{0.8em}
\caption{Analysis on pairwise regularizations.}
\label{tab:ablation2}
\end{table}

% \begin{table}[htb]
% \centering
% \caption{Noise Robustness Results.}
% \begin{tabular}[t]{cc}
% \toprule
% Noise Ratio & $e_{3D}$ \\
% \midrule
%  0\% & 0.043\\
%  1\% & 0.044\\
%  5\% & 0.065\\
%  10\% & 0.083\\
%  20\% & 0.183\\
% \bottomrule
% \end{tabular}
% \label{tab:ab2}
% \end{table}

\section{Conclusion}
We present PR-RRN, a novel deep-networks based approach to NRSfM. We introduce a novel Residual-Recursive Network, which can estimate the 3D shape and camera rotation from 2D inputs. We propose a rigidity-based pairwise contrastive loss and a pairwise consistency loss for regularizing the shape representation learning without assuming global distribution or manifold. Experiments on CMU MOCAP and PASCAL3D+ datasets show that the proposed method achieves state-of-the-art shape recovery accuracy for large-scale human motion and categorical objects reconstruction. PR-RRN is also capable to reconstruct facial landmarks and meshes.

\section*{Acknowledgments}
This work was done when Haitian Zeng interned at Baidu Research. Yuchao Dai was supported in part by National Natural Science Foundation of China (61871325) and National Key Research and Development Program of China (2018AAA0102803). We would like to thank the anonymous reviewers and the area chairs for their useful feedback.

{\small
\bibliographystyle{ieee_fullname}
\bibliography{egbib}
}

\end{document}